\documentclass[twocolumn]{article}
\usepackage{aaai}

\usepackage[ruled,vlined,linesnumbered]{algorithm2e}
\usepackage{amssymb}
\usepackage[super]{nth}
\usepackage{graphicx,amsmath,url}
\usepackage[noend]{algpseudocode}
\usepackage{multirow}
\usepackage{adjustbox}
\usepackage{graphicx} \usepackage{multirow}

  \newtheorem{definition}{Definition}

\usepackage{color-edits}

\newcount\Comments  
\definecolor{darkgreen}{rgb}{0,0.6,0}
\definecolor{orange}{rgb}{1,0.4,0}
\definecolor{purple}{rgb}{1,0,1}
\newcommand{\kibitz}[2]{\ifnum\Comments=1{\textcolor{#1}{#2}}\fi}

\nocopyright

\title{Session Analysis using Plan Recognition}

\author{Reuth Mirsky and Ya'akov (Kobi) Gal \\
Department of Software and Information Systems Engineering \\
Ben-Gurion University of the Negev, Israel \\ 
  \{dekelr,kobig\}@bgu.ac.il\And
  David Tolpin\\
PayPal, Israel\\
dtolpin@paypal.com\\
}

\begin{document}
\maketitle
	
\begin{abstract}
%
  
This paper presents preliminary results of our work with a major financial company, where we try to use methods of plan recognition in order to investigate the interactions of a costumer with the company's online interface. In this paper, we present the first steps of integrating a plan recognition algorithm in a real-world application for detecting and analyzing the interactions of a costumer. It uses a novel approach for plan recognition from bare-bone UI data, which reasons about the plan library at the lowest recognition level in order to define the relevancy of actions in our domain, and then uses it to perform plan recognition. 

We present preliminary results of inference on three different use-cases modeled by domain experts from the company, and show that this approach manages to decrease the overload of information required from an analyst to evaluate a costumer's session --- whether this is a malicious or benign session, whether the intended tasks were completed, and if not --- what actions are expected next.
\end{abstract}

\section{Introduction}
Online websites are often open-ended and flexible interfaces, supporting interaction styles by users that include
  exogenous actions and high noise rate. Such interfaces provide a rich framework for users and are becoming increasingly
  prevalent in many web and mobile applications, but challenge conventional plan
  recognition algorithms for inferring users' activities with the  software.

The open-ended nature of these settings afford a rich spectrum of
interaction for clients: they can perform the same task in many different
ways (such as logging in through a website or by a mobile application),
engage in exploratory activities involving trial-and-error, they
can repeat activities indefinitely (browsing), and they can interleave between
activities. 

This paper focuses on inferring customers' activities for
an Internet financial company, in which customers widely engage
in exploratory behavior, and present initial results for plan
recognition in such settings that can be the basis of future work in this direction.

Our data set consists of customer sessions of a large financial company\footnote{All interactions and examples were altered to preserve privacy}. The customer sessions are recorded as low-level activities of 'click stream' --- web page visits and user interface actions. This data is a rich source of information about customer's behavior, but cannot be used as-is to describe the intentions or plans of the customer: buy product, sell product, manage account, etc. The underlying model used for the recognition process is a set of hierarchical task networks (HTNs) called a ``plan library'', that allows hierarchical decomposition of a task, which can be used both as an explanatory accessory to an overseer and for prediction of future actions.
Our main goal is to detect frauds, validate transactions and predict future actions --- both in terms of the user's intents and in terms of the execution of these intents.

The basic stream of information can be thought of as a recording of customer's
\textit{conversation} with the system. Understanding the language of conversation would bring
benefits in various areas. For example:
\begin{itemize}
\item Predictive verification --- by following customer's actions in
real time, we can have our systems guess next actions and
alert when an agent biases from their predictive plan.
\item Proactive validation --- if we can understand the purpose of
customer's visit based on the beginning of their activity, we
can let our systems start validation of the session even
before it takes place, facilitating faster response and better 
utilization of resources.
\item Computer security --- certain types of fraud are performed by
malicious software controlling customer computers. We may be able to distinguish between malicious and benign actors on the other side of the wire.
\end{itemize}
However, as a flexible interface, the system allows the user to engage in more than one task at a time (making multiple purchases), make mistakes (pressing a button twice instead of once), repeat actions (browse between different parts of the interface) and more.  All of these are features that define an exploratory environment~\cite{reddyGalShieber09}.

While these traits make the interface useful and convenient to the user, it challenges conventional plan
  recognition algorithms for inferring users' activities with the
  software, as they need to reason about all these factors ~\cite{AG13,uzan2015plan}.

\section{Related Work}
The tasks described in the introduction are traditionally addressed by activity recognition algorithms, which are used to tokenize the stream of data into detectable actions~\cite{bao2004activity,liao2006location,Rick2017}. However, these algorithms usually cannot perform higher levels of inference to describe the agent's behavior or predict the future actions of the agent ~\cite{hammerla2016deep,Kristina2017}. Different approaches also used an HMM representation or data-driven learning to elicit tasks from low-level activities~\cite{natarajan2008logical,madani2009prediction}.

Some notable exceptions are works by ~\cite{qin2004attack,duong2005activity,talamadupula2014coordination} which try to combine low level activities and higher level domain knowledge, or ~\cite{talamadupula2014coordination} which propose a multi-agent model for robot collaboration based on plan recognition. However, the works in these lines of research tend to use a domain-theory based recognition or other probabilistic representations, which do not capture the hierarchical nature of task decomposition like Hierarchical task networks (HTNs) and plan libraries.

In order to address the above tasks, we need to use hierarchical plan
recognition, a field of research exploring algorithms that
recognize the plans of the agent based on a partial sequence of
actions, and predict future actions ~\cite{kautz87}.
Few works~\cite{Megenuzzi2017,geib2001plan} did use a plan library as the underlying domain knowledge for the task, but they did not provide predictions or used this information to formalize the output. Other works~\cite{jarvis2005identifying} do output a complete hierarchy, but the plan recognition algorithm used does not work well in exploratory environments.

In this work, we present a criteria for constructing the plan library that will then be used by the agent to perform the plan recognition task. Thus, we do not need to use an activity recognizer. We integrate instead the low-level activities directly into the plan recognition task.

\section {Definitions}
Plan recognition takes as input a plan library and a partial sequence of observations, and outputs a set of possible explanations.

\subsection{Plan Library}
The plan library can be thought of as the language of actions, as well as goals we want to be able to
recognize. The library must specify \textit{terminals}, \textit{non-terminals},
\textit{goals}, and \textit{derivation rules}.

\begin{definition} (Explanation)
\label{def:pl}
A plan library is a tuple $PL=\langle T,NT,G,R \rangle$, where $T$ is a set of are low-level, observable action,$NT$ is a set of complex level actions, composed from
either $T$s or other $NT$s, $G$  is a subset of non-terminals corresponding to the
highest level of actions, representing the goals the agent can
achieve and $R$ is a set of derivation rules which specify how each complex
action can be decomposed into other actions
of the form $ \langle nt \rangle \rightarrow \langle sequence \rangle \mid \langle order \rangle$, where:
\begin{itemize}
\item $\langle nt \rangle$   is a non-terminal, a complex action.
\item $\langle sequence \rangle$ is a sequence of actions which compose the
  complex action.
\item $\langle order \rangle$ is a list of tuples defining the
  the ordering of actions --- actions must appear in 
  the same order as in the tuple, but ordering between different
  tuples is unspecified. For example, `(login, addName)` states
  that the user must log-in before adding a card.
 \end{itemize}
\label{def:plan-library}
\end{definition}

Consider the following very simple plan library for our domain.

\begin{itemize}
\item $T = \{ login, addName, addCredit, signup, \\
submit, home, payment, success, transfer, confirm\}$
\item $NT = \{ AddAccount, Buy \}$
\item $G = \{ AddAccount, Buy \}$
\item $R =  \\
 AddAccount \rightarrow login, addName, addCredit\\
                      \mid [(login,addName) (addName,addCredit)]\\
 AddAccount \rightarrow signup, addName, submit \mid [(signup,addName) (addName,submit)]\\
 Buy \rightarrow home, payment, success \mid [(home,payment) (payment,success)]\\
 Buy \rightarrow home, transfer, confirm \mid [(home,transfer) (transfer,confirm)] $
\end{itemize}

\subsection{Plans and Explanations}
Based on the definitions above, we can apply an algorithm to a
sequence of actions to obtain explanations about ongoing
behaviors and anticipated future actions. 

 A plan is a labeled tree $p=(V,E,\mathcal{L})$, where $V$ and $E$ are the nodes and edges of the tree, respectively,
 and $\mathcal{L}$ is a labeling function $\mathcal{L}: V \rightarrow NT\cup T$ mapping every node in the tree to either a basic or a complex action in the plan library. Each inner node is labeled with a complex action such that its children nodes are a decomposition of its complex action into constituent actions according to one of the refinement methods.
The set of all leaves of a plan $p$ is denoted by $leaves(p)$, and a plan is said to be {\em complete} iff
all its leaf nodes are labeled basic actions, i.e., $\forall v\in leaves(p), \mathcal{L}(v)\in T$.

 An \emph{observation sequence} is an  ordered set of basic actions that represents actions carried out by the observed agent.
 A plan $p$ \emph{describes} an observation sequence $O$ iff every observation is mapped to a leaf in the tree. Formally, there exists an injective function $f:O\rightarrow leaves(p)\cap T$ such that $f(o)=v$.
 The observed agent is assumed to plan by choosing a subset of complex actions as intended goals and then carrying out a separate plan for completing each of these goals.

An agent may pursue several goals at the same time.
Therefore, an explanation can include a set of plans, as described in the following definition:
\begin{definition} (Explanation)
\label{def:exp}
 An explanation for an observation sequence
  is a set of plans such that each plan describes a mutually
  exclusive subset of the observation sequence and taken together the  plans describe all of the observations.  We then say that the explanation \emph{describes} the observation sequence.
\end{definition}

\begin{figure}[t]
 \centering
 \includegraphics[width=8cm]{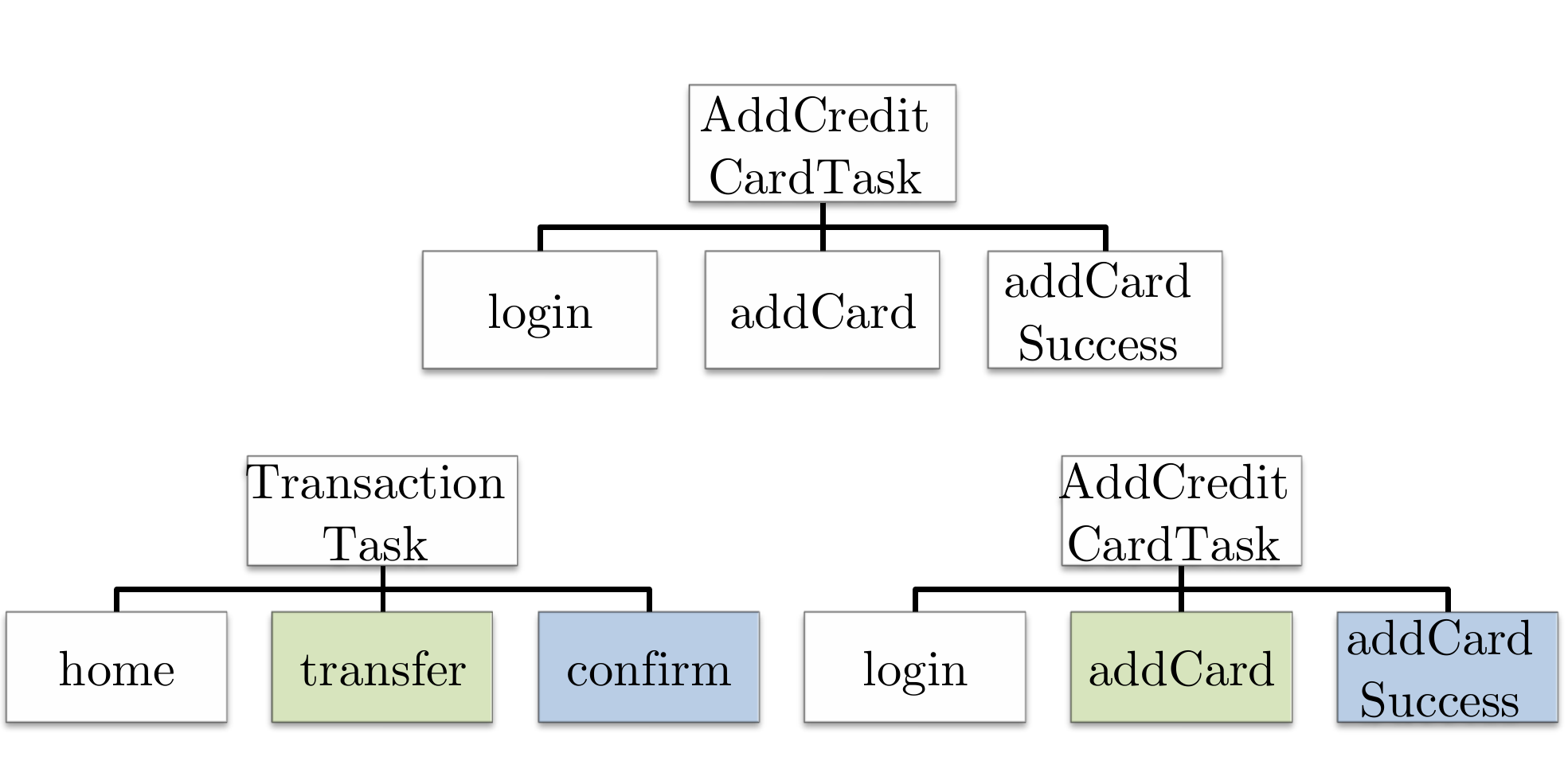}
 \caption{An Explanation with Three Plans.}
 \label{fig:exp}
\end{figure}

For the rest of this paper we will use a running example based on
the following sequence of processed observations:
$[home, login, addName, login, addCredit]$
When we submitted the sequence and the plan library to a plan
recognition algorithm, we obtained two explanations. The first one is presented in Figure~\ref{fig:exp}.

This explanation is a set of trees. In each tree the leaves
are either the actions we observed, or actions predicted 
to perform in the future. A special tag $frontier$ specifies
the location in the tree from where the expansion may 
continue, based on future actions. Leaves with this tags are colored
in green in the figure. Future actions that still has preceding actions
that have not been executed are colored with blue.
If a tree has no `frontier` leaf (top tree in our example), 
then the goal is fully described by actions. 

The output above represents the following explanation
of the input sequence: (1) The agent performed one buy transaction and two account
  additions; (2) The agent performed the first action in the buy task, but
  then stopped -- we expect that the next action to complete this
  task is $transfer$;
 (3) Regarding the account additions, one was completed 
  and the other was not.

For our very basic example, one can think about a brute-force
algorithm for explaining the activity. However in general, a
sophisticated algorithm is required to provide short and
meaningful explanations quickly, in a manner suitable for online
processing. The remaining of this paper talks about the very beginning of work
in collaboration with analysts and developers from the financial company.

\section{Recognition Components}
 \label{sec:Rec}
We now detail the components we use to perform the plan recognition task from the raw click stream of costumer sessions.

\subsection{Preprocessing}
A session is the basic unit of interaction between a customer and the system. Efficient analysis and validation of sessions in real time is crucial for technical realization of the company's business. The raw input in our domain is a stream of costumer sessions. Each session is constructed from a list of entries, where each entry has a specific timestamp, user information and a special label describing the page on the company's site the user is visiting. The average length of a session was $80.49$ entries.

 As a first step, we needed to reduce this number as many entries are irrelevant to the costumer's main tasks in the system and integrating them into terminals in the plan library would be both inefficient and less informative. We defined the following criteria for a session entry to be considered relevant to us:
 \begin{definition} (Landmark in PL)
\label{def:landmark}
A raw entry $e$ in a complete session $S$ is considered a landmark in relation to a plan library $PL$ if $S \setminus \{e\}$ cannot describe a plan from $PL$.
\end{definition}
In collaboration with our partners inside the company, we elicited about 20 types of entries, defined by their page labels, which are landmarks for executing the relevant tasks. We classified each of the pages into a suitable basic action as they appear in our plan library. Counting only these pages as relevant actions we wish to observe, we received an average of $9.68$ observations per session (with stdev of $7.26$).

However, as an exploratory environment, many entries in a session are redundant even if they can be considered landmarks, as the user can make mistakes or perform a landmark action more times than necessary. To perform an elimination of these entries and to remain only with the most relevant observations, we performed another sifting. We discarded all observations that were a repetition of the earlier observation. After this elimination, we reduced the average number of relevant observations per session to $3.68$ (stdev=$2.62$).

\subsection{Recognition}
There are many possible strategies that a costumer can use to perform interactions, and variations within each due to exploratory activities and mistakes carried out by the costumer.
Even after the preprocessing we enforced on the set of actions performed by the costumer, there are still two types of exploratory behaviors which can hinder the ability of the recognizer the infer the costumer's actions correctly: (1) Exogenous actions: even after our preprocessing effort, many actions cannot be combined together when looking at the complete sequence, thus the plan recognizer cannot output any explanation the described all of the actions in the session. (2) One explanation may relate a given action to a relevant task, while another may relate this action to a failed attempt or a mistake. The space of possible explanations can become very large, even for a small number of observations.

The CRADLE algorithm~\cite{Mirsky2017} proposes solutions for both of these problems using three components:
\begin{description}
\item [Inference] it receives as input a plan library, an observation sequence, and outputs a set of explanations, 
each of which is an explanation of the observation sequence in the sense of Definition~\ref{def:exp}. 
\item [Filters] it filters redundant explanations according to a set of
domain-independent conditions. A filter is a function taking a
candidate explanation $e$, and returning \emph{true} or \emph{false}
depending on whether the candidate explanation does or does not pass
a certain condition. For our presented domain, we tried several values and filter types and finally set 3 filters: (1) The number of plans in the explanation is less than or equal to the average; (2) The number of frontier nodes in the explanation is less than or equal to the average; (3) The number of different plans in the explanation is less than 4. We discard all explanations which do not pass these thresholds
\item [Exogenous Actions] CRADLE can handle exogenous actions and mistakes and 
can omit them from the set of  explanations as needed.
\end{description}

\section{Empirical}

We used a click stream of selected sessions, labeled by the company's analysts as sessions containing the relevant tasks. In total, we tested 3 types of sessions, with 50 instances of each session type: Buy, Add account for existing user (AAExist) and Add account for new user (AANew).
For each of these sessions, we performed the process described in Section~\ref{sec:Rec}.

\begin{figure}[t]
 \centering
 \includegraphics[width=7cm]{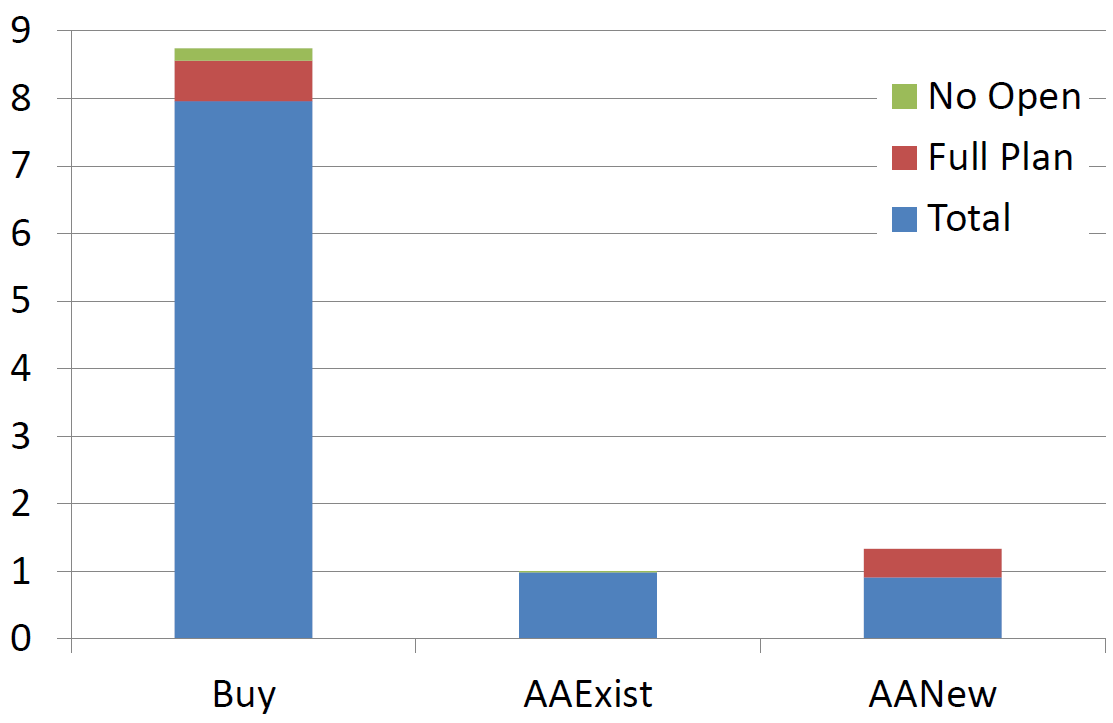}
 \caption{Average Number of Explanations by Type.}
 \label{fig:type}
\end{figure}

Figure~\ref{fig:type} presents the average number of different types of explanations outputted by CRADLE per task: Total is the total number of explanations; Full Plan is the number of explanations in which at least on plan was completed; No Open means the number of explanations in which all plans were completed and had no open frontier. Notice that even in sessions that were labeled by analysts as relevant to some task, the plan recognition process shows that only a small portion of the sessions contain a completed task. These results are similar in the PHATT runs, even though it does not discard explanations, thus might output more explanations with a completed task. We intend to validate these results with other data to evaluate if these sessions were indeed sessions of incomplete tasks.

  In order to evaluate the performance of CRADLE, we compared the runtime and number of outputted explanation in comparison to the PHATT algorithm, augmented with the exogenous actions handling of CRADLE (without this augmentation, PHATT would not be able to output any explanation at all that describes the complete sequence of observations).
  Table~\ref{tab:results} summarizes these runs.

\begin{table}[b]
\centering
\label{tab:results}
\begin{tabular}{|c|r|r|r|}
\hline
                   & \multicolumn{1}{c|}{Buy} & \multicolumn{1}{c|}{AAExist} & \multicolumn{1}{c|}{AANew} \\ \hline
Session Entries    & 147.02                   & 14.13                                & 80.33                              \\ \hline
Observations       & 5.64                     & 2.06                                 & 3.33                             \\ \hline
CRADLE Explanations & 8.74                     & 1.00                                 & 1.33                               \\ \hline
PHATT Explanations & 18.32                     & 1.00                                 & 2.00                               \\ \hline
CRADLE Time (seconds)      & 0.06                     & 0.01                                 & 0.01                               \\ \hline
PHATT Time (seconds)         & 0.07                     & 0.01                                 & 0.03                               \\ \hline
\end{tabular}
\caption{Runtime and Explanation Set Size for CRADLE and PHATT}
\end{table}

The first and most important thing to notice in the table, is the difference between the original length of a session entry, and the outputted set of explanations by the plan recognizers. An average decrease of $83\%$ in the number of entities representing the session. Such a decrease allows a faster analysis of the session, since the number of possible explanations is exponential with the number of observations~\cite{Mirsky2017}. For example, even with the pruning of CRADLE, the average number of explanations it outputted for the AANew sessions is $146.06$ (compared to $1.33$ with the preprocessing). Moreover, the outputted explanations have a structure imposed by the plan library that the original stream lacked.

The second point to notice it that the values of the AAExist case are similar in both algorithms. We attribute this to the fact that performing this task is always performed in the same fashion, without any options to bias in the interface of the company's website.

\section{Future Work}
This a preliminary work for using plan recognition for exploratory environments on real-world click streams. It uses a novel approach for plan recognition from bare-bone UI data, which reasons about the plan library in the lowest recognition level in order to define the relevancy of actions in our domain, and then uses it to perform plan recognition.

While we manage to process low level data using mostly tools intended for higher levels of inference, there is still much to be done from both ends of the process:
First, we expect that we can use most intelligent tools for the preprocessing stage, either from the world of activity recognition or natural language processing. As we started to show here, we wish to use the domain knowledge (e.g. the plan library) in our low-level action detection.
Second, we wish to visualize this information in a coherent manner that will allow analysts evaluate sessions in real time for fast verification and validation, or to be able to counter adversarial behavior in time.

%


\bibliography{libb}
\bibliographystyle{alpha}

\end{document}